\documentclass[10pt,twocolumn,letterpaper]{article}

\usepackage{iccv}
\usepackage{times}
\usepackage{epsfig}
\usepackage{graphicx}
\usepackage{amsmath}
\usepackage{amssymb}
\usepackage{stfloats}  
\usepackage[linesnumbered,ruled,lined]{algorithm2e}
\usepackage{booktabs}
\usepackage{pifont}
\usepackage{bbding}
\usepackage{subfigure}

\usepackage[pagebackref=true,breaklinks=true,letterpaper=true,colorlinks,bookmarks=false]{hyperref}

 \iccvfinalcopy 

\def\iccvPaperID{8470} 

\ificcvfinal\fi

\begin{document}

\title{Spatio-Temporal Matching for Siamese Visual Tracking}

\author{Jinpu Zhang \qquad
Yuehuan Wang \\
School of Artificial Intelligence and Automation,  \\
Huazhong University of Science and Technology, Wuhan 430074, China\\
{\tt\small zjphust@gmail.com}	\qquad
{\tt\small yuehwang@hust.edu.cn}
}

\maketitle
\ificcvfinal\thispagestyle{empty}\fi

\begin{abstract}
   Similarity matching is a core operation in Siamese trackers. Most Siamese trackers carry out similarity learning via cross correlation that originates from the image matching field. However, unlike 2-D image matching, the matching network in object tracking requires 4-D information (height, width, channel and time). Cross correlation neglects the information from channel and time dimensions, and thus produces ambiguous matching. This paper proposes a spatio-temporal matching process to thoroughly explore the capability of 4-D matching in space (height, width and channel) and time. In spatial matching, we introduce a space-variant channel-guided correlation (SVC-Corr) to recalibrate channel-wise feature responses for each spatial location, which can guide the generation of the target-aware matching features. In temporal matching, we investigate the time-domain context relations of the target and the background and develop an aberrance repressed module (ARM). By restricting the abrupt alteration in the interframe response maps, our ARM can clearly suppress aberrances and thus enables more robust and accurate object tracking. Furthermore, a novel anchor-free tracking framework is presented to accommodate these innovations. Experiments on challenging benchmarks including OTB100, VOT2018, VOT2020, GOT-10k, and LaSOT demonstrate the state-of-the-art performance of the proposed method. 
   
\end{abstract}

\section{Introduction}

\begin{figure}[t]
\centering
\includegraphics[width=\linewidth]{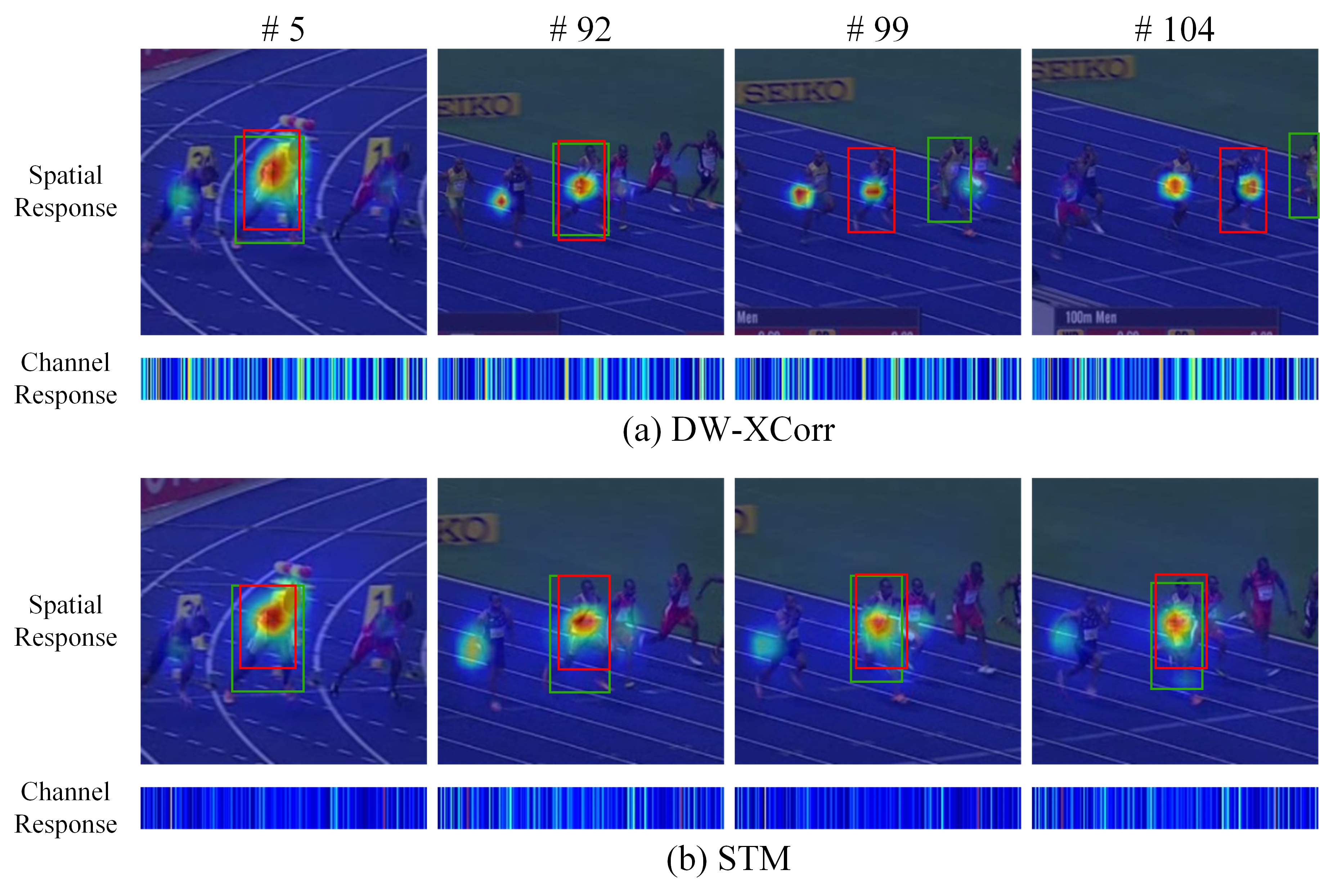}
	\caption{Comparison of DW-XCorr and STM. STM indicates spatio-temporal matching. The spatial responses are generated by averaging all channels of each pixel, the channel responses are generated by normalizing the maximum of each channel. The green box and red box represent groundtruth and prediction respectively. Our STM considering the matching of channel and time dimensions is more robust to distractors.}
\label{fig:fig1}
\end{figure}

Visual object tracking is a fundamental task in computer vision. It aims to infer the location of a target in subsequent frames of the video based on the target state in the first frame. Object tracking has been widely adopted in the field of surveillance\cite{surveillance}, robotics\cite{robotics}, autonomous driving\cite{autonomous} and human-computer interaction\cite{human}, etc. Despite its rapid progress in recent decades, challenges such as occlusion, deformation and background interference\cite{otb13,otb15} still must be overcome.

Recently, Siamese network\cite{siam} based trackers have attracted attention due to their favorable balance of speed and tracking performance. These trackers formulate the tracking problem as learning a general similarity metric between the feature embedding of the target template and the search region. Therefore, how to embed the information of the two branches to obtain informative response maps is particularly vital. The seminal work, SiamFC\cite{SiamFC}, utilizes cross correlation (XCorr) for similarity measurement.
Furthermore, SiamRPN\cite{siamrpn} extends up-channel cross correlation (UP-XCorr) to adapt to the RPN structure, however, this gives rise to an imbalanced parameter distribution and hinders training optimization. To address this issue,  SiamRPN++\cite{siamrpn++} presents a lightweight depth-wise cross correlation (DW-XCorr) to achieve efficient information association. Almost all of the current Siamese trackers\cite{siamfc++,siamcar,siamban,ocean} employ DW-XCorr for similarity matching. Some recent approaches\cite{cgacd,AR} of local pixel-wise correlation are also studied by researchers.

Despite the great success of the current cross correlation, it still has several limitations. First, cross correlation originates from image matching\cite{match}. The core concept is to combine feature maps using the inner product and evaluate the similarity on each subwindow. Siamese network only provides an efficient way to implement this operation with convolution. It is essentially still a 2-D spatial matching and assigns the same weight to the matching results of each channel of deep features. In practice, the data-driven deep features are sparsely activated\cite{tadt}. When applied to the tracking task, only a few channels are active in describing the target. The remaining large portion of channels that contain redundant and irrelevant information cannot benefit the matching. We calculate the normalized maximum values of each channel, as shown in Figure \ref{fig:fig1}a. The channel response of DW-XCorr is dense since both the target and distractors are activated. This makes it difficult for the tracker to discriminate the desired target from distractors. The channel attention mechanism\cite{se, bam, cbam} is usually used for channel-wise feature calibration in the classification task. However, it squeezes global spatial information into a single channel descriptor and hence emphasizes the features of all semantic categories. This is not appropriate for the tracking task that needs to distinguish instances.

The second limitation of traditional correlation is the neglect of temporal context in tracking. Siamese trackers decompose the tracking problem into independent matching problems for each frame. Such approaches are prone to fail in case of e.g., occlusion, fast deformation or presence of distractors, where a single-frame matching model is insufficient for robust tracking. If we can utilize the previous frame and construct temporal constraint relations, these aberrances are easily repressed.
To this end, traditional trackers\cite{moose,KCF,HCF} implement a simple linear update strategy for the template. However, the constant learning rate on all spatial dimensions across all frames is often inadequate to cope with changing updating requirements in different tracking scenarios\cite{updatenet}. Another approach is learning an online classifier\cite{atom,dimp} from the tracking sequence. Such a model is more discriminative but at the cost of time-consuming iterative optimization.

\subsection{Contributions}
In view of the aforementioned issues, we leverage the 4-D information of height, width, channel and time for comprehensive matching, called Spatio-Temporal Matching.

In spatial matching, we increase the exploration of channel-wise matching and design a space-variant channel-guided correlation (SVC-Corr) that can emphasize target-aware responses and weaken interference responses. 
The innovation of SVC-Corr is two-fold:   
1) It establishes various channel associations between the template and each subwindow in the search area. This allows an independent and accurate recalibration of the response to a specific instance at each location without semantic disturbance from other spatial locations.
2) It is a learnable correlation module that can benefit from large-scale offline training, rather than a handcrafted parameter-free metric such as cross correlation. 
Figure \ref{fig:fig1}b shows the sparse channel responses using SVC-Corr. The generated target-aware responses are more effective in separating semantic objects than the DW-XCorr features.

In temporal matching, we propose an aberrance repressed module (ARM) to investigate the hidden information propagated in the interframe response maps. Specifically, an efficient optimization function is added to restrict the abrupt alteration of the interframe response maps in both training and testing stages. It takes into account the variation of target and background simultaneously in the time domain, and adjust the response map flexibly for different tracking scenarios. 
As shown in Figure \ref{fig:fig1}b, the heatmap of \textit{frame 5} imposes a strong constraint on the subsequent heatmaps, thus avoiding the drifting. Importantly, our ARM  brings negligible additional computation cost.

Using the introduced SVC-Corr and ARM, we present a novel anchor-free tracking framework, termed Siamese Spatio-Temporal Matching  (SiamSTM) tracking network. The effectiveness of the framework is verified on five benchmarks: OTB100\cite{otb15}, VOT2018\cite{vot2018}, VOT2020\cite{vot2020}, GOT-10k\cite{got10k} and LaSOT\cite{lasot}. Our approach achieves leading performance, while running at 66 FPS.

\section{Related Work}

Siamese-based trackers have attracted wide attention due to their superior performance and speed. Similarity matching is one of their most crucial components. The pioneering method SiamFC\cite{SiamFC} utilizes XCorr to obtain a single channel response map and the instance with the highest similarity score is considered as the target. XCorr efficiently realizes the similarity measurement of traditional image matching in the form of convolution. 
Following this similarity-learning work, CFNet\cite{CFNet} incorporates the Correlation Filter into the Siamese network for end-to-end training. DSiam\cite{dsiam} adds a fast transformation learning that enables online learning of target appearance variation and background suppression from previous frames. Dong\cite{siamtri} employs triplet loss\cite{triplet} to take full advantage of the relationship among the exemplar, positive instance and negative instance instead of the simple pairwise relationship. Although these methods improve the matching ability to some extent, they still leverage XCorr for information embedding.

SiamRPN combines Siamese network with the region proposal network (RPN)\cite{Faster}, which discards the traditional multi-scale tests. To embed the information of anchors, XCorr is extended by adding a huge convolutional layer to scale the channels (UP-XCorr). However, the heavy up-channel module leads to a severe imbalance of the parameter distribution. SiamRPN++ presents a lightweight DW-XCorr to efficiently generate a multi-channel correlation response map. Thus, the parameters on the template and the search branches are balanced, making the training procedure more stable. Since RPN-based Siamese trackers are sensitive to hyper-parameters associated with the anchors, later works focus on anchor-free models, such as SiamFC++\cite{siamfc++}, SiamCAR\cite{siamcar}, SiamBAN\cite{siamban} and Ocean\cite{ocean}. They directly predict classification and regression results in a per-pixel-prediction manner, and still use DW-XCorr to encode similarity.

Several recent studies\cite{cgacd,AR} state that the part-level relations are more robust to variations than global matching in tracking and propose pixel-wise correlation to implement this idea. However, similar to DW-XCorr, this matching method ignores the influence of channel association that is one of the main factors for generating target-aware features as discussed above. The pixel-to-global matching in PG-Net\cite{PGNet} adds channel kernels on this basis to unify the local positions similarity of the various channels.

All of the above are single-frame independent matching regardless of temporal information. The main use of temporal information in the current methods is updating the template with simple linear interpolation\cite{moose,KCF,HCF,dai} or learning an online classifier\cite{atom,dimp,d3s,drol}. The former have been proved insufficient in most tracking situations\cite{updatenet}, while the latter takes excess time on the frames that require gradient optimization. By contrast, this paper extends the idea of matching to the time domain and achieves both high robustness and high efficiency.

\section{Method}

\begin{figure*}[t]
\centering
\includegraphics[width=0.9\linewidth]{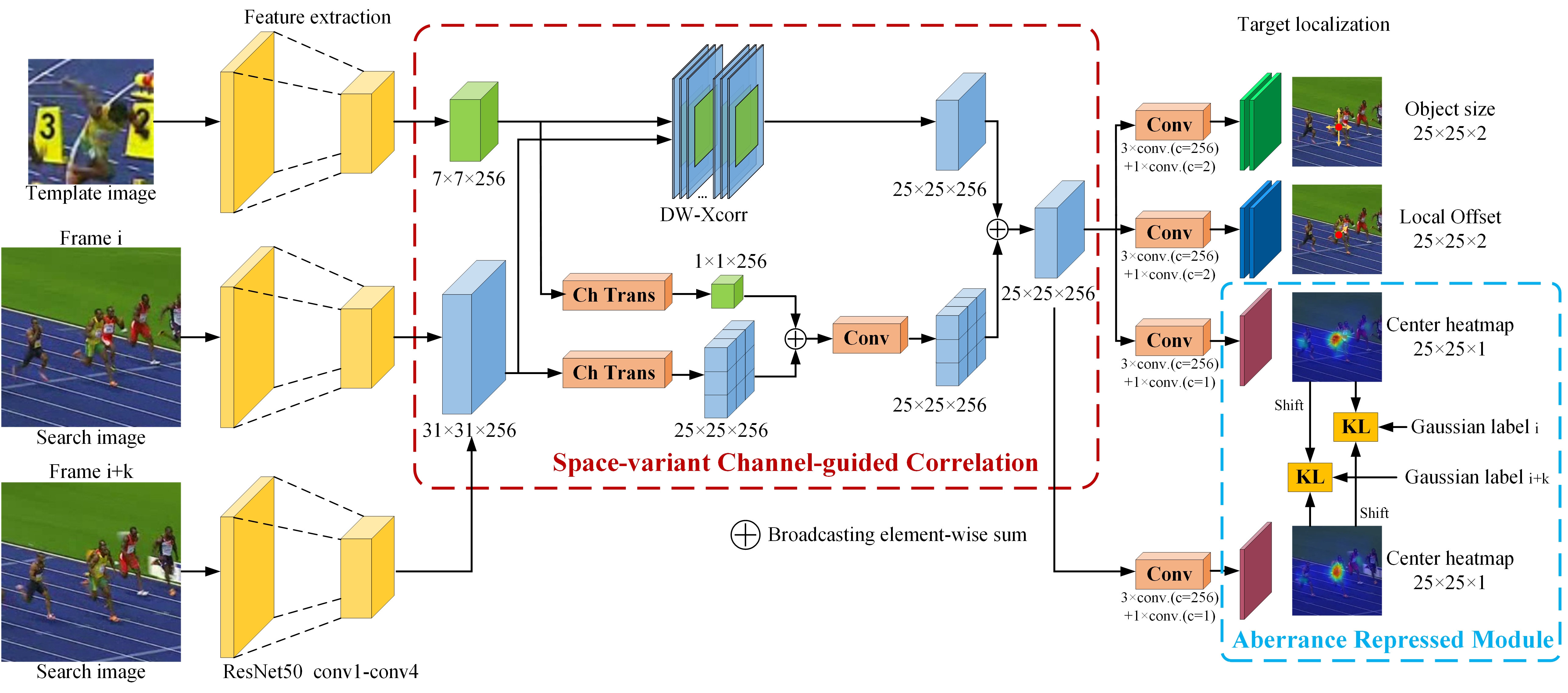}
   \caption{Overview of the proposed network. The features of the template and the search region are first extracted by a Siamese subnetwork. Then, SVC-Corr is exploited for spatial matching. The generated response map is fed into the tracking head subnetwork for target localization. Among the outputs, the center heatmap performs temporal matching across multiple frames using ARM. }
\label{fig:overview}
\vspace{-0.2cm} 
\end{figure*}

This section details the proposed SiamSTM. Figure \ref{fig:overview} illustrates an overview of the framework. In the following, we first introduce the basic Siamese tracker that is inspired by CenterNet\cite{centernet}. Then we elaborate the designed Spatio-Temporal Matching, which consists of a space-variant channel-guided correlation for spatial matching and an aberrance repressed module for temporal matching.

\subsection{Basic Siamese Tracker}
\label{base}

The basic Siamese tracker is composed of feature extraction subnetwork and target localization subnetwork, as shown in Figure \ref{fig:overview}. In the feature extraction subnetwork, we modify the ResNet50\cite{resnet} according to SiamRPN++\cite{siamrpn++} to make it more suitable for the tracking task. Moreover, we cut off the $ conv5 $ of the ResNet50 and only retain the $ conv1 - conv4 $ to reduce computations.

In the target localization subnetwork, considering that anchor-based trackers rely on a large number of heuristic hyper-parameters\cite{siamfc++} and lack the competence to amend the weak predictions\cite{ocean}, we leverage an anchor-free structure. The right part of Figure \ref{fig:overview} illustrates its architecture that represents the object with object center, object size and local offset\cite{centernet}. Let $ \hat{Y}\in \mathbb{R}^{H \times W \times 1}  $ be an output center heatmap of height $ H $ and width $  W $. For an arbitrary position $ (x,y) $ in the heatmap, a prediction $ \hat{Y}_{x,y}=1 $ corresponds to the tracking object center, while $ \hat{Y}_{x,y}=0 $ is the background.
The classification label $ Y \in \mathbb{R}^{H \times W \times 1} $ adopts a Gaussian function $ Y_{x,y}=exp\left ( -\frac{\left ( x-x_{c}\right )^{2}+ \left ( y- y_{c}\right )^{2}}{2\sigma _{p}^{2}}\right ) $, where $  ( x_{c},y_{c} )  $ denotes the coordinate of the target center point in the heatmap, and $ \sigma _{p} $ is an object size-adaptive standard deviation\cite{cornernet}. The training objective is a penalty-reduced pixel-wise logistic regression with focal loss\cite{FL}. We refer readers to \cite{cornernet, centernet} for more details.

To eliminate the discretization gap caused by the spatial stride  $ R=8 $, an additional local offset $ \hat{O}\in \mathbb{R}^{H \times W \times 2}  $ is calculated for each center point.
For the object size, we directly predict a $ \hat{S}\in \mathbb{R}^{H \times W \times 2}  $  to represent the height and width in each location.
The supervisions of local offset and object size only act at the unique target location in the tracking task, and other locations are ignored. In particular, the offset label $ \{o_{x},o_{y}\} $ is $ \left \{ \frac{i}{R}- \lfloor \frac{i}{R} \rfloor, \frac{j}{R}- \lfloor \frac{j}{R} \rfloor \right \} $ for the target position $ (i,j) $ in the original image, and the size label $ \{s_{w},s_{h}\} $ is the groundtruth width and height after downsampling. Both of these are trained with L1 loss. The overall training objective of the basic tracker is:
\begin{align}
L_{base} = L_{cls} + \lambda_{off}L_{off} +  \lambda_{size}L_{size}
\label{con:loss_base}
\end{align}
where $L_{off}, L_{size}  $ are the loss functions of local offset and object size. $ \lambda_{off}, \lambda_{size} $ are the trade-off hyper-parameters.

We argue that our center-based structure is more suitable for object tracking than the currently popular FCOS-based structure. The Gaussian label-assign method in the center-based structure is equivalent to unifying the classification score and localization quality in the FCOS-based structure into a joint and single representation. Thus, it eliminates the training-test inconsistency mentioned in \cite{gfocal} and enables a stronger correlation between classification and localization quality.
Additionally, the ideal prediction heatmap should be a Gaussian-like probability distribution, and we can easily constrain its alteration in the time domain as detailed in Sec. \ref{ARM}.

\subsection{Space-variant Channel-guided Correlation}

\begin{figure}[t]
\centering
\includegraphics[width=0.95\linewidth]{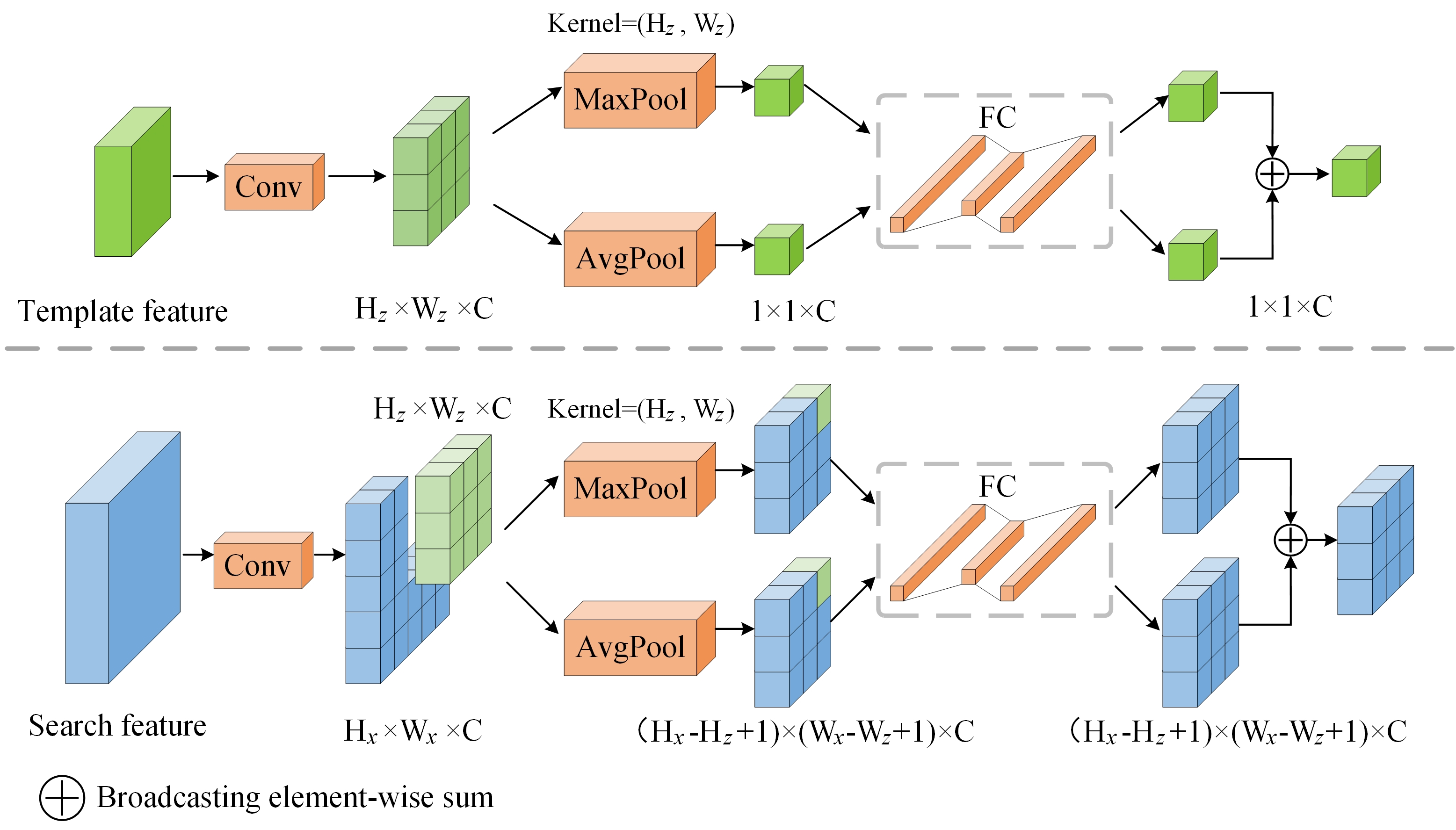}
	\caption{The architecture of Ch Trans. It utilizes max pooling outputs and average pooling outputs with a shared network. The pooling kernel size is equal to the template feature size. }
\label{fig:ch_trans}
\vspace{-0.3cm} 
\end{figure}
 
The difference between 2-D image matching and 3-D deep feature matching, namely modeling of the associations between the channels, has been analyzed above. The convolutional network learns specific semantic information mainly based on a few channels. When applied to the tracking task, we can obtain the target-aware feature responses by identifying channels that are active to the target and are inactive to the interference.

For this purpose, we introduce a space-variant channel-guided correlation (SVC-Corr) to establish channel correspondence between the template and the search area on the basis of traditional 2-D correlation. The red dotted box in Figure \ref{fig:overview} illustrates its structure, including two parallel branches. Given the template feature $ z \in  \mathbb{R}^{H_{z} \times W_{z} \times C}  $ and the search feature  $ x \in  \mathbb{R}^{H_{x} \times W_{x} \times C}  $, we first employ a DW-XCorr to obtain a spatial response, i.e., $ f_{sa}(z,x)=z \star x $, where $ \star $ denotes DW-XCorr and $ f_{sa}(z,x) \in  \mathbb{R}^{(H_{x}-H_{z}+1) \times (W_{x}-W_{z}+1) \times C} $ is the correlation result.

The DW-XCorr branch embeds the template feature into the search features along height and width, while the other branch aims to channel-wise embedding. The core step is the extraction of channel descriptors for template and search area, meanwhile modeling the interdependencies between the channels. This is implemented by the Channel Transform (Ch Trans) in Figure \ref{fig:ch_trans}. 
Specifically, after a $ 3\times 3 $ convolution $ \varphi_{1} $, we squeeze the spatial dimension of the feature through max pooling $ f_{k}^{max} $ and average pooling $ f_{k}^{avg} $ to gather different clues about the channel importance\cite{cbam}. Note that the pooling kernel size $ k $ is equal to the template feature size (i.e., $ f_{k}^{max, avg}(z) \in \mathbb{R}^{1 \times 1 \times C}, f_{k}^{max, avg}(x) \in \mathbb{R}^{(H_{x}-H_{z}+1) \times (W_{x}-W_{z}+1) \times C} $), hence we can obtain various channel descriptors from each search subwindow (\textit{a sliding window of the same size as template feature size in the search region}) and associate them with the template respectively. The pooled descriptors are further fed into a shared multiple fully connected ($ FC $) layer to capture the channel-wise interdependencies.  Considering that there are multiple channel descriptors on the search feature, we can efficiently implement all $ FC $ operations in parallel with $ 1 \times 1 $ convolution. The learned descriptors are merged by element-wise summation. In short, the Ch Trans is computed as:
\begin{align}
T_{ch}(\omega)= FC(f_{k}^{max} (\varphi_{1}(\omega))) \oplus  FC(f_{k}^{avg} (\varphi_{1}(\omega))) 
\label{con:ch_trans}
\end{align}%
where $ \omega \in \{z,x\} $ denotes template or search features, $ \oplus  $ is element-wise summation.

With the channel descriptors of the template and search area, we adopt broadcasting element-wise summation followed by $ 1 \times 1 $ convolution $ \varphi_{2} $ to generate space-variant weights for different channels at each location. Finally, these weights recalibrate channel-wise feature responses of the previous DW-XCorr using element-wise summation. The whole process of SVC-Corr can be expressed as:
\begin{align}
f_{ca}(z,x) &= \varphi_{2}(T_{ch}(z) \oplus T_{ch}(x)) \\
svc\_corr(z, x) &= f_{ca}(z,x) \oplus  f_{sa}(z,x)
\label{con:cacorr}
\end{align}%
where $ f_{ca}  $  and $ ca\_corr $ denote the channel weights and SVC-Corr result, respectively.

\begin{figure}[t]
\centering
\includegraphics[width=0.95\linewidth]{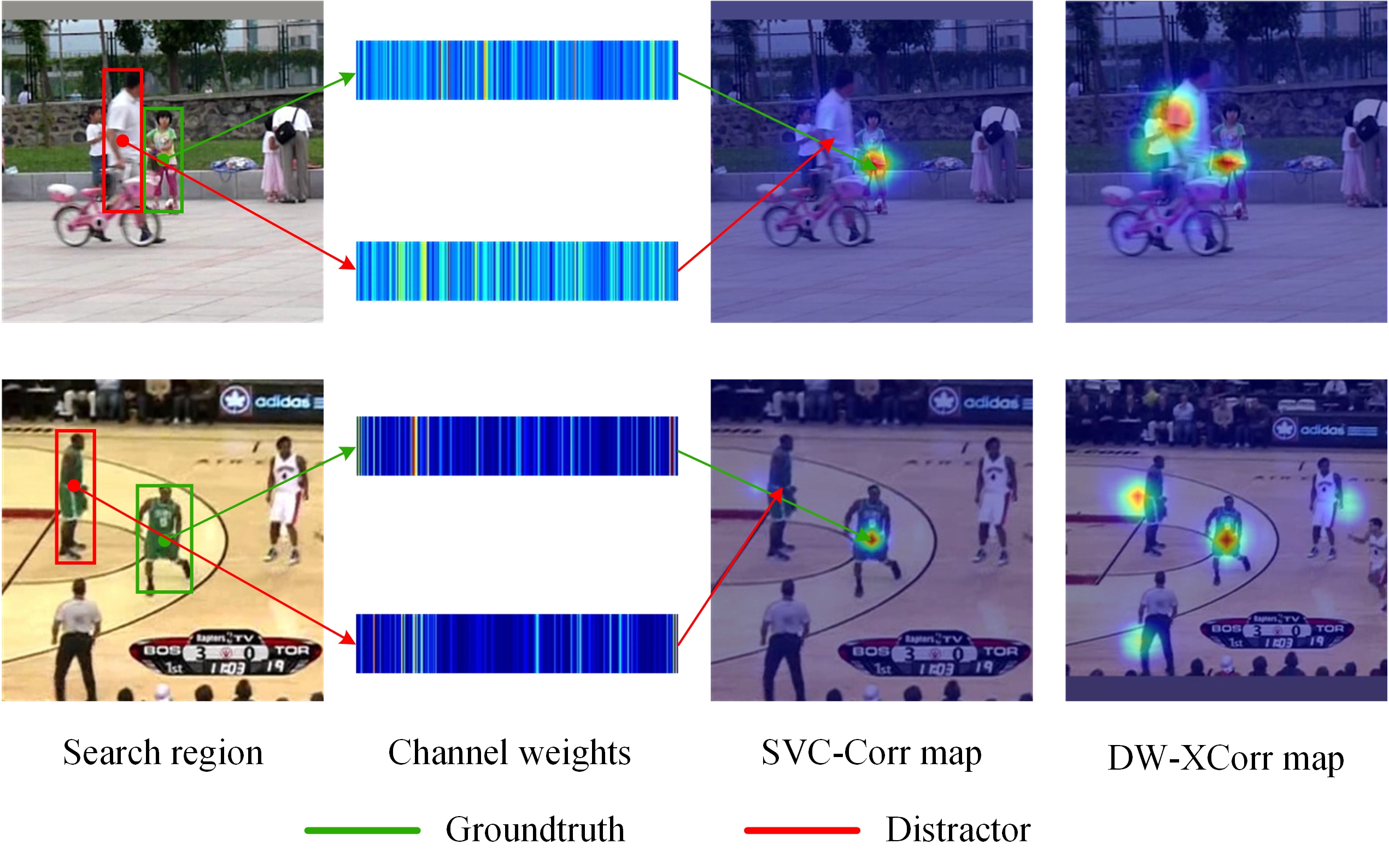}
	\caption{Comparison of SVC-Corr and DW-XCorr. The second column represents the channel weights $ f_{ca}  $ on the green target area and red distractor area. Due to the space-variant channel associations between the template and each search subwindow, SVC-Corr maps (third column) can focus on the target regions. }
\label{fig:ca_dw}
\vspace{-0.3cm} 
\end{figure}

Our SVC-Corr has two prominent advantages. 1) Unlike channel attention that focuses on all semantic categories, our SVC-Corr is tailored to different instances. For each subwindow in the search area, SVC-Corr associates it with the template independently in channel dimension, i.e., $ f_{ca}  $ has a total of $  (H_{x}-H_{z}+1) \times (W_{x}-W_{z}+1) $ space-variant channel weights. This makes the channel information of various spatial locations not interfere with each other, and thus is more beneficial for the generation of discriminative target-aware responses. Figure \ref{fig:ca_dw} shows a comparison between SVC-Corr and DW-XCorr. 2) All parameters in Eq. \eqref{con:ch_trans}- \eqref{con:cacorr} are learnable rather than handcrafted and can benefit from the large-scale offline training.

\subsection{Aberrance Repressed Module}
\label{ARM}

\begin{figure}[t]
\centering
\includegraphics[width=0.95\linewidth]{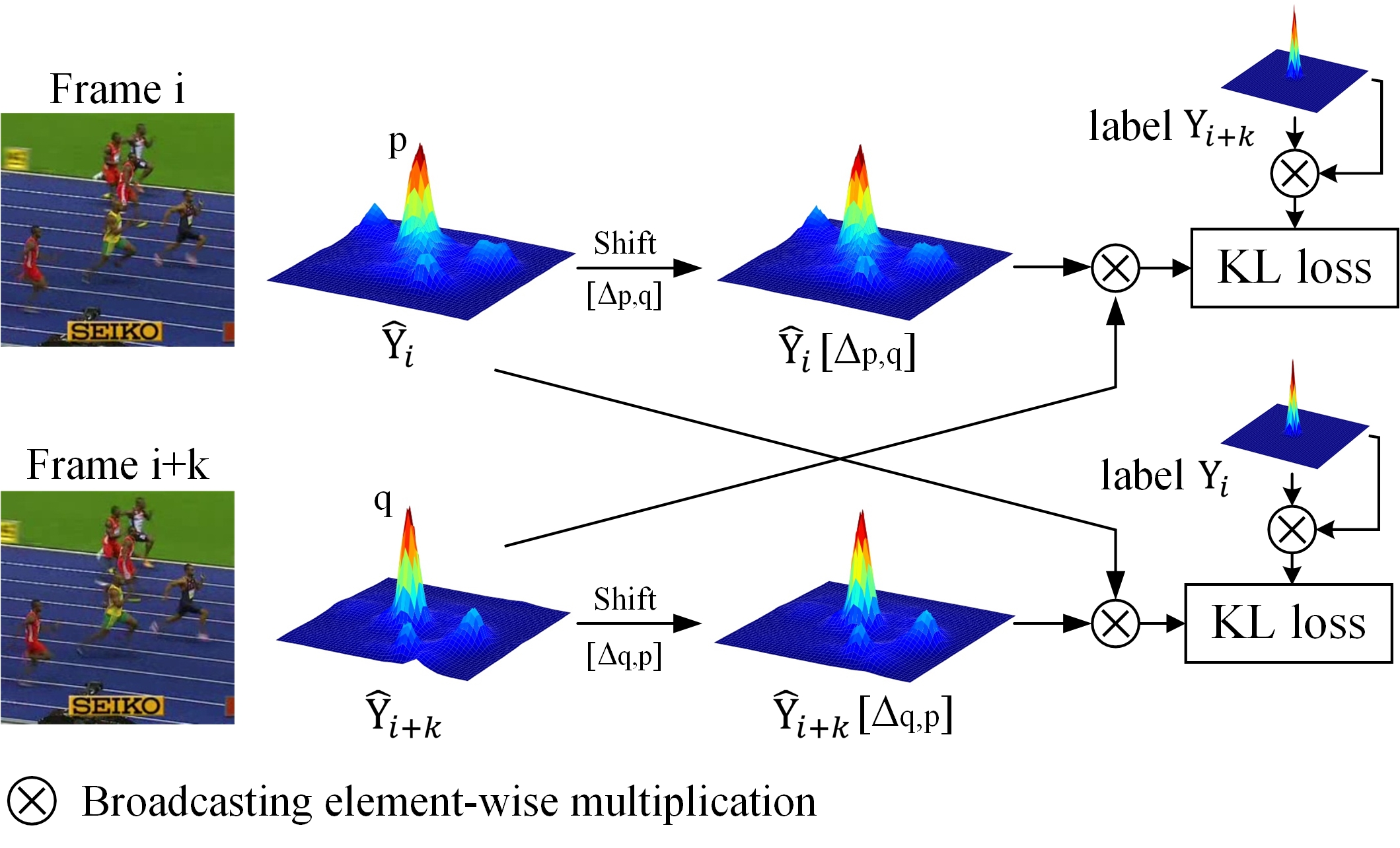}
	\caption{The architecture of ARM. It restricts the alteration of the input heatmaps in the time domain by symmetric KL loss.}
\label{fig:arm}
\vspace{-0.3cm} 
\end{figure}

After fully exploiting the 3-D information of deep features for single-frame matching, we further extend matching to time-dimension to explore the information hidden across multiple frames. Figure \ref{fig:arm} indicates our temporal matching method, aberrance repressed module (ARM).

Instead of raw image patches, center heatmaps of the adjacent frames are used as the input data for our ARM. Based on the above analysis and existing works\cite{arcf, autotrack}, we observe two criteria: 1) An ideal heatmap should be unimodal with narrow peak. 2) The distribution of heatmaps between the adjacent frames should be as similar as possible after alignment. Consequently, the purpose of ARM is to maximize the similarity of the peak-aligned heatmaps between the adjacent frames and minimize their errors with Gaussian labels. We propose to train the objective function with KL divergence:
\begin{align}
\begin{split}
L_{arm}= & KL\left ( Y_{i+k} \otimes  Y_{i+k}, \,  \hat{Y}_{i}[\Delta_{p,q}] \otimes  \hat{Y}_{i+k} \right)  \\
		+  &  KL\left ( Y_{i} \otimes  Y_{i}, \,  \hat{Y}_{i+k}[\Delta_{q,p}] \otimes  \hat{Y}_{i} \right)  
\end{split}
\label{con:kl}
\end{align}
where $ \hat{Y}_{i}, \hat{Y}_{i+k}, Y_{i}, Y_{i+k}  $ denote heatmaps and Gaussian labels for adjacent frame $ i $ and $ i+k $, as detailed in Sec. \ref{base}. $ p,q $ represent peak locations of $ \hat{Y}_{i}, \hat{Y}_{i+k} $, and $ \Delta_{p,q}  $ indicates the circular shifting operation to coincide peak $ p $ with peak $ q $, while $ \Delta_{q,p}  $ indicates the opposite. $ \otimes $ is element-wise multiplication. The KL divergence is calculated as $  KL(y, x) = \int y \log(y) - \int  ylog(x) $.

In the first line of Eq. \ref{con:kl}, the $  \hat{Y}_{i}[\Delta_{p,q}] \otimes  \hat{Y}_{i+k}  $ term fuses the two peak-aligned heatmaps, the distribution of which can reflect the alteration between two frames. A sharp unimodal distribution indicates that the two heatmaps are similar, and the loss between it and the Gaussian label is low. When an aberrance occurs, this distribution will change suddenly, e.g., the peak value decreases or additional interference peaks appear, causing a high loss. Therefore, the minimization of this loss function can restrict the alteration of heatmaps to suppress aberrances. The second line of Eq. \ref{con:kl} is symmetrical with the first line, but the Gaussian label is replaced by that of another frame. This makes the most of the monitoring information of each frame and leads to a more reliable restriction.

The ARM realizes the temporal matching without extra parameters and can be trained end-to-end in conjunction with the above basic tracker. The overall loss function is:
\begin{align}
L= L_{base}+\lambda_{arm}L_{arm}
\label{con:loss_all}
\end{align}
where $ L_{base} $ comes from the basic tracker loss in Eq. \ref{con:loss_base} and $ \lambda_{arm} $ is a hyper-parameter.  

\begin{algorithm}[tb] 
\caption{SiamSTM online tracking}  
\label{alg:online}
\LinesNumbered  
\KwIn{Sequence of length L, Initial bounding box}
\KwOut{Target bounding box for each frame }
\textbf{Initialize:}  $ t=1 $ \\
heatmap $ \hat{Y}_{1} $ $  \leftarrow$ basic tracker with SVC-Corr \\
label $ Y $, peak position $ p $  $  \leftarrow$ initial box \\
Update $ Y_{last} \leftarrow Y, \hat{Y}_{last} \leftarrow \hat{Y}_{1}, p_{last} \leftarrow p $ \\
\For{$t \in [2,L]$}{
	$ \hat{Y}_{t},   \hat{O}_{t}, \hat{S}_{t}  $ $  \leftarrow$ basic tracker with SVC-Corr  \;
	Top $ K $ response positions $  \{q_{1},..., q_{K}\} $ in $ \hat{Y}_{t},  $ \;
	\For{$k \in [1,K]$}{
		Shift $ Y_{last}, \hat{Y}_{last} $ according to $ \Delta_{p_{last}, q_{k}} $ \;
		Calculate $ L_{arm}^{k} $ according to Eq.\ref{con:kl} \;
	}
	$ \hat{k}  \leftarrow argmin_{k} L_{arm}^{k}$  \;
	\If{$ \hat{k} \neq 1 $}{
		$ \hat{Y}_{t} \leftarrow \left(1+ \hat{Y}_{last}[ \Delta_{p_{last}, q_{\hat{k}}}] \right) \otimes \hat{Y}_{t} $ \;
		Update $ Y_{last} \leftarrow Y_{last}[ \Delta_{p_{last}, q_{\hat{k}}}], \hat{Y}_{last} \leftarrow \hat{Y}_{last}[ \Delta_{p_{last}, q_{\hat{k}}}], p_{last} \leftarrow q_{\hat{k}} $ \;
	}
	Hanning window and scale penalty\cite{siamrpn} on $ \hat{Y}_{t} $ \;
	$ \hat{x}_{t}, \hat{y}_{t} \leftarrow argmax_{x,y} \hat{Y}_{t}$ \;
	$ (\delta\hat{x}_{t}, \delta\hat{y}_{t}) \leftarrow \hat{O}_{\hat{x}_{t}, \hat{y}_{t}},
	   (\hat{w}_{t}, \hat{h}_{t}) \leftarrow \hat{S}_{\hat{x}_{t}, \hat{y}_{t}} $ \;
	Output $ box_{t} \leftarrow \left( \hat{x}_{t}+\delta\hat{x}_{t}, \hat{y}_{t}+\delta\hat{y}_{t}, \hat{w}_{t}, \hat{h}_{t} \right) $
}
\end{algorithm}

The online tracking procedure is summarised in Algorithm \ref{alg:online}. After initialization, the basic tracker with SVC-Corr first generates center heatmap $ \hat{Y}_{t} $, offset prediction $ \hat{O}_{t} $ and size prediction $  \hat{S}_{t} $ (line 6). Lines 7-16 indicate the inference process of ARM. Specifically, we align the label $ Y_{last} $ and the heatmap $ \hat{Y}_{last} $ of the previous frame with the top $ K $ response locations of $ \hat{Y}_{t} $ respectively, and then calculate the KL divergence $ L_{arm}^{k} $  for each pair. Note that only the first term in Eq. \ref{con:kl} is calculated here, since we only have the label of the previous frame. The $ q_{\hat{k}} $ that minimize $ L_{arm}^{k} $ can be considered as the optimal peak position with the temporal constraint. If $ q_{\hat{k}} $  is different from the original peak position $ q_{1} $ in $ \hat{Y}_{t} $, we add $ \hat{Y}_{last}[ \Delta_{p_{last}, q_{\hat{k}}}] $ as a weight on  $ \hat{Y}_{t} $ (line 14) and update related variables  (line 15). Finally, the maximum response location on the weighted heatmap $ \hat{Y}_{t} $ is extracted as the target center (line 18), combining with the offset prediction and the size prediction on this point to produce the bounding box (lines 19-20). 
Our ARM utilizes temporal information during both raining and inference. Compared with linear updating, it benefits from large-scale offline training and considers both the target and the background. Consequently, ARM is more flexible in adjusting each position of the response for different tracking situations. 
Meanwhile, it is much more efficient than online classifiers and achieves a tracking speed of 66 FPS.

\section{Experiments}

This section presents the results of SiamSTM on OTB100\cite{otb15}, VOT2018\cite{vot2018}, VOT2020\cite{vot2020}, GOT-10k\cite{got10k} and LaSOT\cite{lasot}, with comparisons to the state-of-the-art algorithms. A detailed ablation study is then performed to evaluate the effects of each component in our model.

\subsection{Implementation Details}

The proposed SiamSTM is trained on ImageNet VID\cite{Imagenet}, ImageNet DET\cite{Imagenet}, Youtube-BB\cite{ytb}, COCO\cite{coco} and GOT-10k\cite{got10k} for experiments on OTB100\cite{otb15}, VOT2018\cite{vot2018} and VOT2020\cite{vot2020}. In addition, for experiments on GOT-10k\cite{got10k} and LaSOT\cite{lasot}, the model is only trained with their official training set for a fair comparison. To train ARM, the training input is a triplet containing a template image and two search images with an interval less than 100 frames.
The template size is $ 127\times127 $ pixels, while the search image is $ 255 \times 255 $ pixels. The training batch size is set to 96 and a total of 20 epochs are trained with SGD on 4 TITANV. We use a learning rate that linearly increased from 0.001 to 0.005 for the first 5 warmup epochs and then exponentially decayed to 0.0005 for the rest of 15 epochs. For the first 10 epochs, we freeze the parameters of the backbone. For the remaining epochs, the backbone network is unfrozen, and the whole network is trained end-to-end. The hyper-parameters $ \lambda_{off},  \lambda_{size} $ and $  \lambda_{arm} $ in Eq. \ref{con:loss_base} and Eq. \ref{con:loss_all} are set to 1, 0.1 and 0.5, respectively. The $ K $ in Algorithm \ref{alg:online} is set to 3.

\subsection{State-of-the-art Comparison}
\begin{table*}[t]\small\centering
\newcommand{\tabincell}[2]{\begin{tabular}{@{}#1@{}}#2\end{tabular}}
\begin{tabular}{@{}ccccccccccc@{}}
\toprule
    & \tabincell{c}{LADCF \\ \cite{ladcf}} & \tabincell{c}{UPDT \\ \cite{updt}}  & \tabincell{c}{SiamRPN++ \\ \cite{siamrpn++}} & \tabincell{c}{SiamFC++ \\ \cite{siamfc++}} & \tabincell{c}{SiamBAN \\ \cite{siamban}} & \tabincell{c}{SiamMask \\ \cite{siammask}} & \tabincell{c}{SiamRCNN \\ \cite{siamrcnn}} & \tabincell{c}{DiMP  \\ \cite{dimp}}  & \tabincell{c}{PGNet \\ \cite{PGNet}} & SiamSTM \\ \midrule
EAO $ \uparrow $ & 0.389 & 0.379 & 0.416     & 0.430     & \textcolor{blue}{0.452}   & 0.383    & 0.408    & 0.440  & \textcolor{green}{0.451} & \textcolor{red}{0.488}   \\
A $ \uparrow $   & 0.503 & 0.530  & 0.596     & 0.581    & 0.592   & 0.604    & \textcolor{blue}{0.609}    & 0.597 & \textcolor{red}{0.618} & \textcolor{green}{0.604}   \\
R $ \downarrow $  & \textcolor{green}{0.159} & 0.182 & 0.234     & 0.183    & 0.178   & 0.276    & 0.220     & \textcolor{blue}{0.153} & 0.192 & \textcolor{red}{0.103}   \\ \bottomrule
\end{tabular}
\caption{Performance comparisons on VOT2018. \textcolor{red}{Red}, \textcolor{blue}{blue} and \textcolor{green}{green} fonts indicate the top-3 trackers.}
\label{tab:vot18}
\vspace{-0.1cm} 
\end{table*}

\begin{table*}[!t]\small\centering
\newcommand{\tabincell}[2]{\begin{tabular}{@{}#1@{}}#2\end{tabular}}
\begin{tabular}{@{}ccccccccccc@{}}
\toprule
    & \tabincell{c}{SiamFC \\ \cite{SiamFC}} & \tabincell{c}{SiamRPN++ \\ \cite{siamrpn++}} & \tabincell{c}{SiamRCNN \\ \cite{siamrcnn}} & \tabincell{c}{SiamMargin \\ \cite{vot2019}} & \tabincell{c}{FCOT \\ \cite{FCOT}}  &  \tabincell{c}{ATOM \\ \cite{atom}}  & \tabincell{c}{DiMP \\ \cite{dimp}}  & \tabincell{c}{SiamMask \\ \cite{siammask}} & SiamSTM & \tabincell{c}{SiamSTM \\ mask} \\ \midrule
EAO $ \uparrow $ & 0.179  & 0.247     & 0.233    & 0.226      & 0.247 & 0.271 & 0.281 & \textcolor{blue}{0.328}    & \textcolor{green}{0.299}   & \textcolor{red}{0.391} \\
A $ \uparrow $   & 0.418  & 0.435     & 0.458    & 0.415      & 0.421 & 0.462 & 0.448 & \textcolor{blue}{0.589}    & \textcolor{green}{0.480}    & \textcolor{red}{0.625} \\
R $ \uparrow $   & 0.502  & 0.670      & 0.610     & 0.637      & 0.703 & 0.734 & \textcolor{blue}{0.743} & 0.634    & \textcolor{red}{0.756}   & \textcolor{red}{0.756} \\
M   & $ \times $      & $ \times $         & $ \times $        & $ \times $          & $ \times $     & $ \times $ & $ \times $   & $ \checkmark $       & $ \times $      & $ \checkmark $     \\ \bottomrule
\end{tabular}
\caption{Performance comparisons on VOT2020. "M" denotes whether the mask results are predicted. The last column indicates that the mask branch of SiamMask is added to our SiamSTM to output the object mask. }
\label{tab:vot20}
\vspace{-0.1cm} 
\end{table*}

\begin{table*}[!t]\small\centering
\newcommand{\tabincell}[2]{\begin{tabular}{@{}#1@{}}#2\end{tabular}}
\begin{tabular}{@{}ccccccccccc@{}}
\toprule
     & \tabincell{c}{ECO \\ \cite{eco}}   & \tabincell{c}{SiamFC \\ \cite{SiamFC}} & \tabincell{c}{SiamRPN++ \\ \cite{siamrpn++}} & \tabincell{c}{SiamFC++ \\ \cite{siamfc++}} & \tabincell{c}{ROAM \\ \cite{roam}} & \tabincell{c}{SiamCAR \\ \cite{siamcar}} & \tabincell{c}{ATOM \\ \cite{atom}} & \tabincell{c}{DiMP \\ \cite{dimp}}  & \tabincell{c}{Ocean \\ \cite{ocean}} & SiamSTM \\ \midrule
AO $ \uparrow $   & 0.316 & 0.348  & 0.518     & 0.595    & 0.465 & 0.569   & 0.556 & \textcolor{blue}{0.611} & \textcolor{blue}{0.611} & \textcolor{red}{0.624}   \\
$ \rm SR_{50} $ $ \uparrow $ & 0.309 & 0.353  & 0.618     & 0.695    & 0.532 & 0.670    & 0.635 & \textcolor{green}{0.717} & \textcolor{blue}{0.721} & \textcolor{red}{0.730}    \\
$ \rm SR_{75} $ $ \uparrow $ & 0.111 & 0.098  & 0.325     & \textcolor{green}{0.479}    & 0.236 & 0.415   & 0.402 & \textcolor{blue}{0.492} & 0.473 & \textcolor{red}{0.503}   \\ \bottomrule
\end{tabular}
\caption{Performance comparisons on the GOT-10k test-set.}
\label{tab:got10k}
\vspace{-0.3cm} 
\end{table*}


To extensively evaluate the proposed method, we compare it with state-of-the-art trackers on five challenging tracking datasets. The selected comparison methods include XCorr based tracker\cite{SiamFC, CFNet}, DW-XCorr based trackers\cite{siamrpn++, siammask, siamfc++, siamcar, siamban, ocean}, pixel-wise correlation based tracker\cite{PGNet}, correlation filter based trackers\cite{ladcf, updt, eco}, meta-learning based trackers\cite{atom, dimp, FCOT, roam}, multi-domain learning based trackers\cite{mdnet, vital} and others\cite{vot2019, siamrcnn}.

\noindent\textbf{VOT2018}\cite{vot2018} 
VOT2018 consists of 60 sequences with various challenging factors. The overall performance of the tracker is evaluated in terms of Expected Average Overlap (EAO) that takes both accuracy (A) and robustness (R) into account. Table \ref{tab:vot18} shows the comparison with the existing top-performing trackers on VOT2018. Our SiamSTM attains the best EAO and robustness among all of the methods. Compared with the offline trackers, i.e., anchor-based SiamRPN++\cite{siamrpn++} and anchor-free SiamBAN\cite{siamban}, our model achieves EAO improvements of 7.2 and 3.6 points. It is worth noting that the improvements mainly come from the robustness, obtaining 56\% and 42.1\% increases over SiamRPN++ and SiamBAN, respectively. Impressively, the robustness of our method exceeds those methods relying on online adaptation (LADCF\cite{ladcf}, UPDT\cite{updt}, ATOM\cite{atom} and DiMP\cite{dimp}). This further demonstrates that SiamSTM is more feasible in the use of temporal information. PGNet is superior in terms of accuracy because of its more detailed local pixel-wise correlation, but it is still inferior to our method with respect to robustness and EAO.

\begin{figure}[t]
\centering
\includegraphics[width=0.9\linewidth]{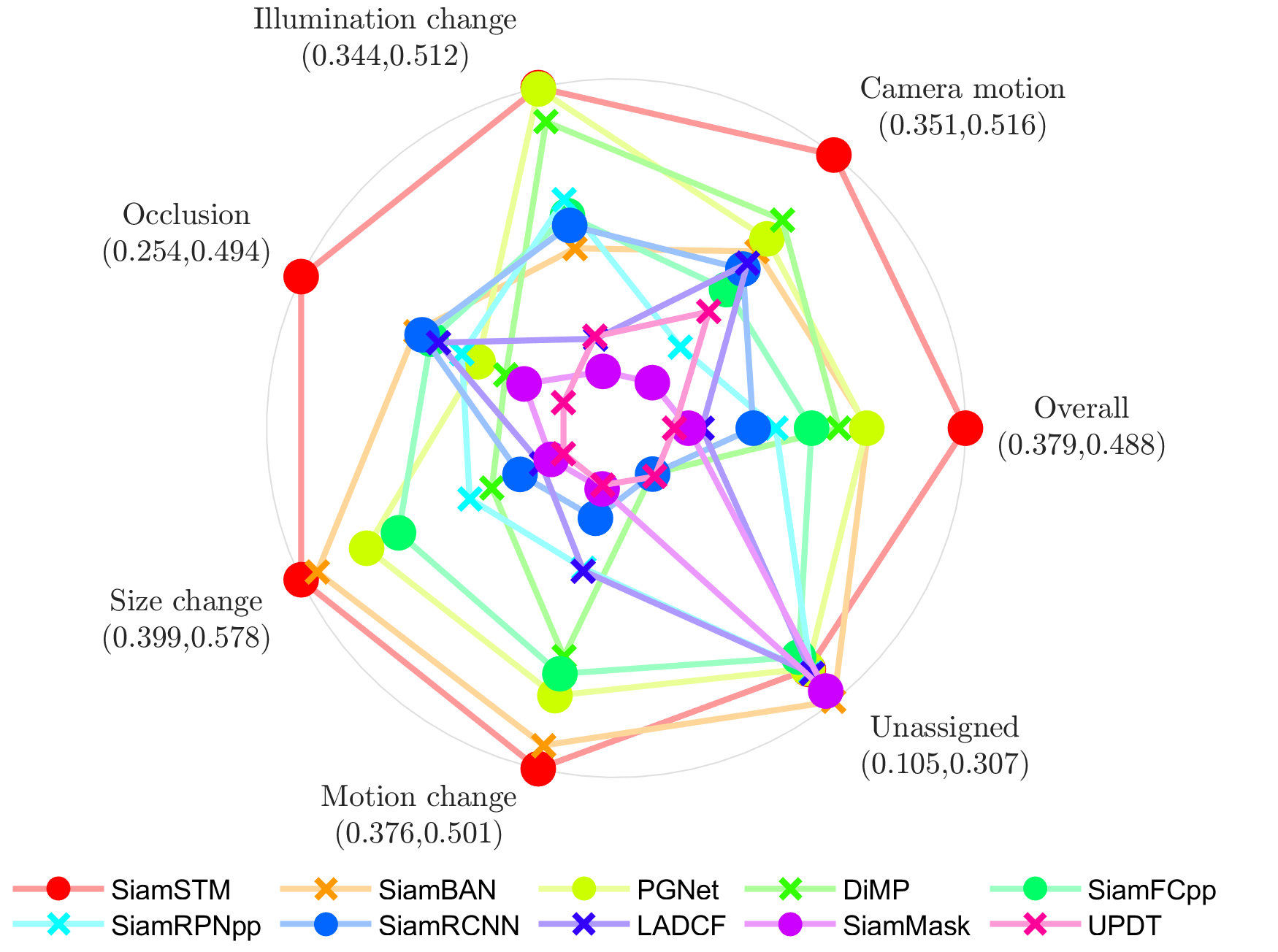}
	\caption{Comparison of EAO on VOT2018 for different visual attributes. The values in parentheses indicate the EAO range of each attribute and the overall EAO range of the trackers.}
\label{fig:attr_eao18}
\vspace{-0.3cm} 
\end{figure}

Sequences in VOT2018 are annotated by the following visual attributes: occlusion, illumination change, motion change, size change and camera motion. Frames that do not correspond to any of the five attributes are denoted as unassigned. We compare the EAO of the above methods on these visual attributes in Figure \ref{fig:attr_eao18}. Our SiamSTM ranks first on almost all of the attributes except for unassigned and outperforms the second place by a wide margin in the challenging occlusion and camera motion.

\noindent\textbf{VOT2020}\cite{vot2020} 
VOT2020 contains 60 sequences and redefines accuracy, robustness and EAO in the context of the new anchor-based evaluation protocol\cite{vot2020}. Another significant novelty is that the target position was encoded by a segmentation mask. Since mask prediction is not our main purpose, we utilize the \textit{bbox} evaluation mode to compare different trackers. Table \ref{tab:vot20} shows the evaluation results on VOT2020, and 'M' denotes whether mask results are predicted. The proposed model achieves the best performance (EAO, A and R) among all of the compared trackers that output the bounding box. The robustness of our model surpasses that of DiMP with elaborate meta-learning by 1.3 points. Moreover, the online mechanism in Algorithm \ref{alg:online} is a linear operation and consumes much less time than the fine-tuning of meta-learning. We observe that SiamMask surpasses SiamRPN++ by more than 8.1 points on EAO and 15.4 points on accuracy. Therefore, we simply complement our framework with the mask branch in SiamMask to obtain the object mask and improve the EAO by 9.2 points and accuracy by 14.5 points. We have reason to believe that  the results can be further improved by a careful modification of the mask prediction.

\noindent\textbf{GOT-10k}\cite{got10k} 
GOT-10k is a large-scale dataset containing over 10 thousand videos and is challenging in terms of zero-class-overlap between the provided train-set and test-set. For a fair comparison, we follow the the protocol of GOT-10k and only train our model with its train-set and evaluate the performance on the test-set of 180 videos. The performance indicators are average overlap (AO) and success rate (SR). As shown in Table \ref{tab:got10k}, the proposed SiamSTM achieves a best AO of 0.624, outperforming previous \textit{SOTA} DiMP\cite{dimp} and Ocean\cite{ocean} by 2.1 points.

\begin{figure}[t] 
  \centering 
    \includegraphics[width=0.46\linewidth]{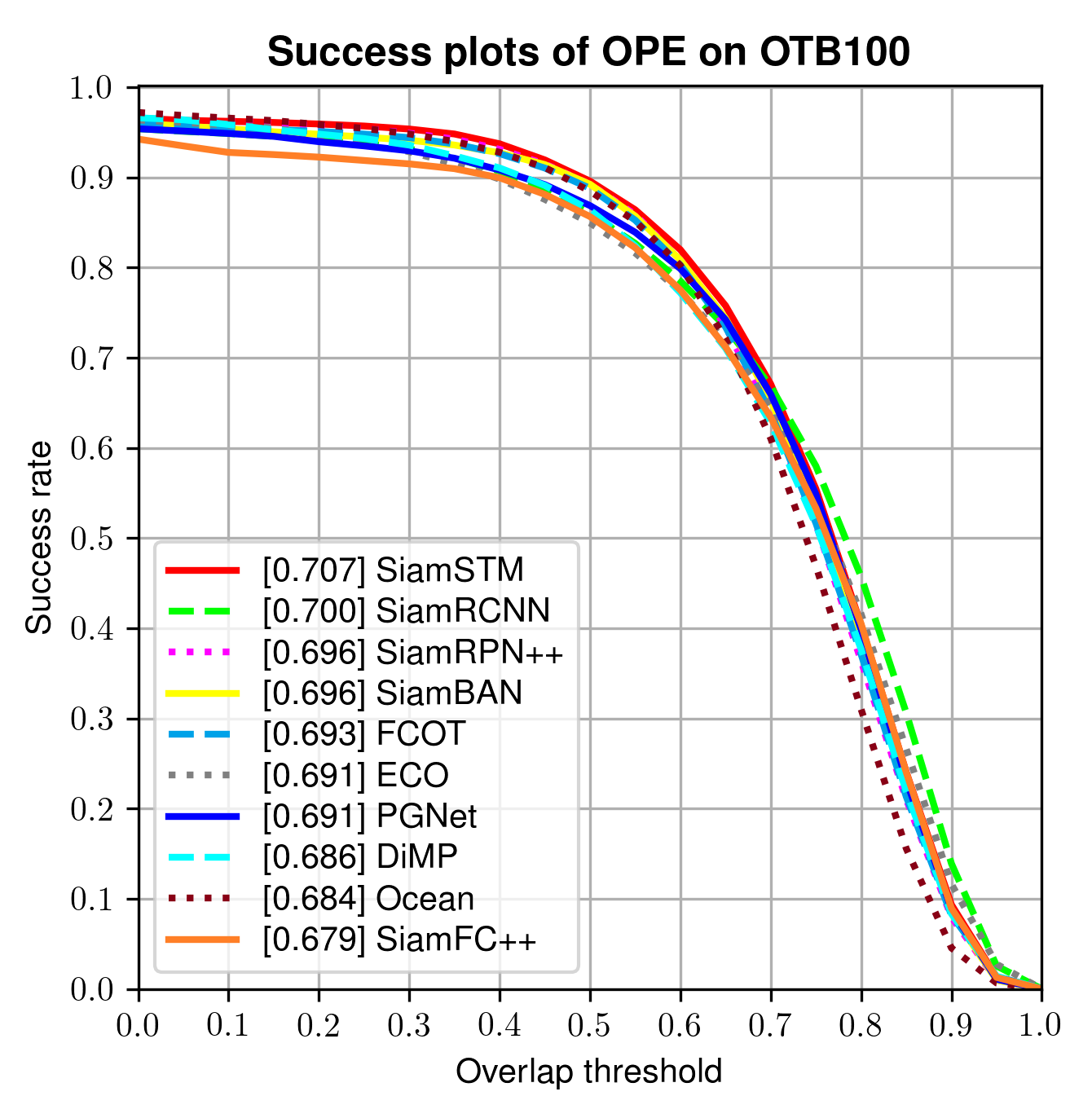} 
    \includegraphics[width=0.46\linewidth]{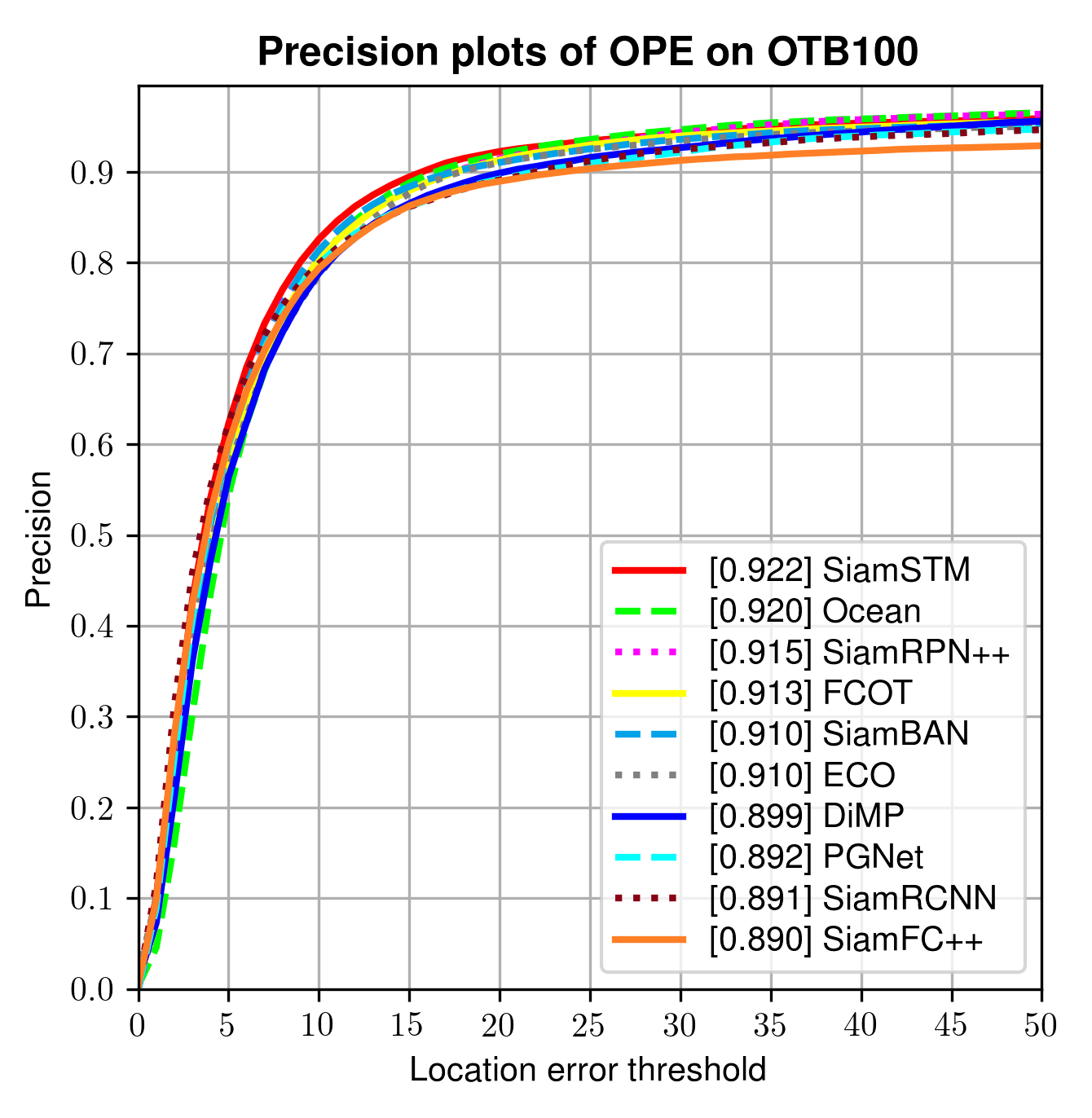} 
\caption{Success and precision plots on OTB100} 
\label{fig:otb} 
\vspace{-0.1cm} 
\end{figure}

\noindent\textbf{OTB100}\cite{otb15} 
OTB100 is a widely used public tracking benchmark consisting of 100 sequences. Precision (Prec.) and area under curve (AUC) are used to rank the trackers. As reported in Figure \ref{fig:otb}, among the compared methods, our SiamSTM achieves the best performance on both precision (0.922) and AUC (0.707). It even improves compared to the redetection-based SiamRCNN\cite{siamrcnn} by 1\% on AUC with a much simple network architecture.

\begin{figure}[t]
  \centering 
    \includegraphics[width=0.46\linewidth]{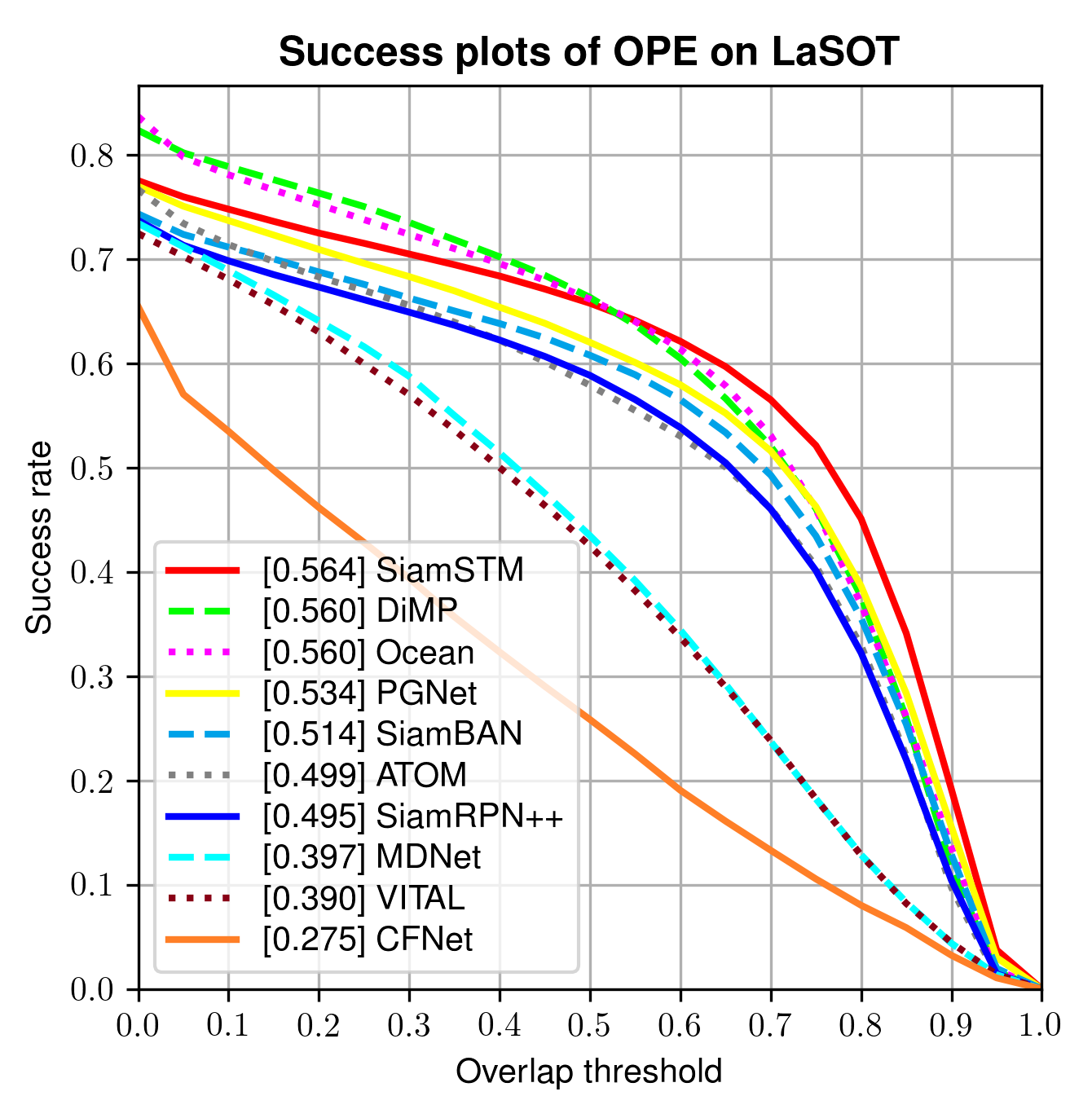} 
    \includegraphics[width=0.46\linewidth]{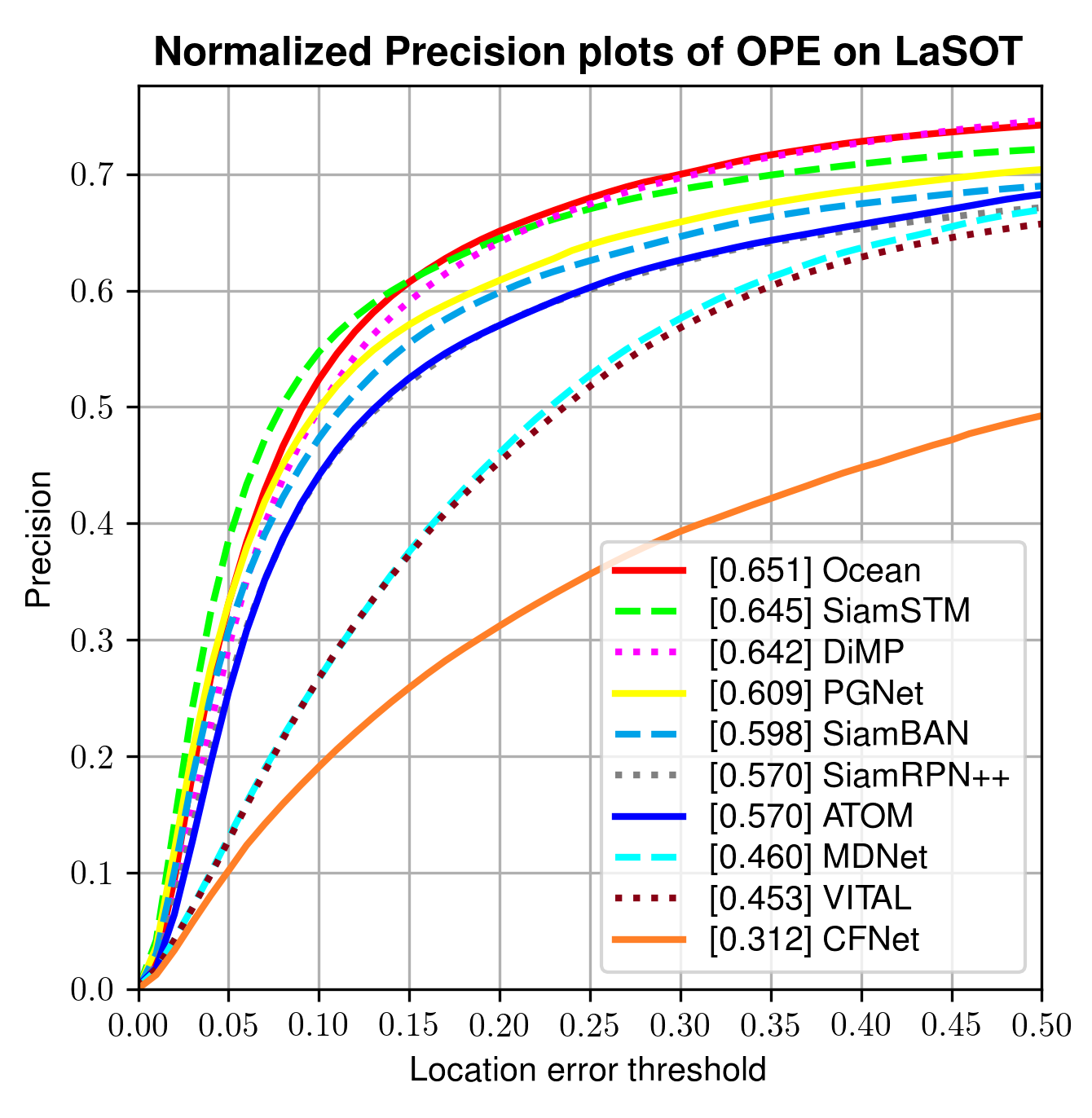} 
\caption{Success and normalized precision plots on LaSOT} 
\label{fig:lasot} 
\vspace{-0.3cm} 
\end{figure}

\noindent\textbf{LaSOT}\cite{lasot} 
To further evaluate the proposed model on long-term tracking, we report the results on LaSOT. Compared with the previous datasets, LaSOT has longer sequences with an average sequence length of more than 2,500 frames. In Figure \ref{fig:lasot}, we show normalized precision and success plots for 280 videos in the test-set. Our tracker ranks first in terms of AUC, second in terms of normalized precision and 0.5\% higher than DiMP. This demonstrates that the 4-D spatio-temporal matching can adapt to the complexity of long-term tracking and avoid model drift.

\subsection{Ablation Study}

To further verify the efficacy of the proposed method, we perform an ablation study on the GOT-10k test-set, as presented in Table \ref{tab:ablation}. Note that triplet input is only required when training ARM, otherwise the training input is still a pair of a template image and a search image.

\noindent\textbf{Head Structure.} 
The FCOS head is used to estimate the distances from each pixel within the target object to the four sides of the groundtruth bounding box, while the Center head is used to estimate the target center point and its corresponding size. By replacing FCOS head with Center head, the AO is improved by 1.7 points (\ding{173} \textit{vs.} \ding{172}). The improvement mainly originates from the difference in the label assign, where our Gaussian label can jointly optimize the classification branch and the quality estimation branch, and thus produces more confident and reliable predictions.

\noindent\textbf{SVC-Corr vs. DW-XCorr.} 
To analyze the contribution of the proposed SVC-Corr, we conduct a comparison to the popular DW-XCorr. The SVC-Corr obtains significant AO gains of 1.6 points compared to DW-XCorr (\ding{174} \textit{vs.} \ding{173}). We attribute these gains to the benefits from the recalibration of the channel-wise feature responses. Moreover, Figure \ref{fig:fig1} and \ref{fig:ca_dw} visualize the responses of these two correlation methods. Our SVC-Corr only activates a few target-aware channels, and meanwhile weaken the remaining redundant and irrelevant channels, hence its activated channel responses are sparser than that of DW-XCorr in Figure \ref{fig:fig1}. Figure \ref{fig:ca_dw} further demonstrates that the SVC-Corr establishes different channel associations between each subwindow and template, making the response maps more capable to discriminate the targets and disturbances. We quantitatively analyze the benefits of this space-variant association, meanwhile integrate it into two other trackers (SiamRPN++\cite{siamrpn++} and SiamBAN\cite{siamban}) to show the generalization ability. Detailed results are provided in the supplementary material.

\noindent\textbf{Aberrance Repressed Module.} 
Last, we evaluate the impact of the temporal matching carried out by ARM. The ARM brings 2.3 points AO gains on DW-XCorr (\ding{175} \textit{vs.} \ding{173}) and 1.8 points AO gains on SVC-Corr (\ding{176} \textit{vs.} \ding{174}), respectively. Additionally, we state that the performance gains benefit from the temporal constraint rather than triplet training input, as shown in the supplementary material. The results verify that ARM can utilize information hidden in the time domain to suppress aberrances and is thus more robust and accurate for object tracking. Figure \ref{fig:fig1}b intuitively proves that ARM is competent in dealing with distractors. 

\begin{table}[t] 
 \centering
\begin{tabular}{@{}c|cccc|c@{}}
\toprule
Num & Head & DW-XCorr & SVC-Corr & ARM & AO    \\ \midrule
\ding{172}     & FCOS       & $ \checkmark $   &     &     & 0.570 \\
\ding{173}     & Center     & $ \checkmark $   &     &     & 0.587 \\
\ding{174}     & Center     &          & $ \checkmark $ &     & 0.603 \\
\ding{175}     & Center     & $ \checkmark $   &     & $ \checkmark $ & 0.610 \\
\ding{176}     & Center     &          & $ \checkmark $ & $ \checkmark $  & \textbf{0.624} \\ \bottomrule
\end{tabular}
\caption{Ablation study on GOT-10k. Head represents the tracking head using FCOS-based structure or Center-based structure.}
\label{tab:ablation}
\vspace{-0.3cm} 
\end{table}

\section{Conclusion}

We propose a novel Siamese Spatial-Temporal Matching (SiamSTM) tracking network based upon the observation that traditional cross correlation ignores the matching relationships of channel and time. Our SVC-Corr realizes the space-variant channel-wise information recalibration for each matching position and yields a target-aware response map. Moreover, by extending temporal matching ARM, our approach efficiently mines the information propagated in multiframes to suppress the aberrances during tracking. Both innovations can benefit from large-scale offline training and run with a high speed. Comprehensive experiments on five benchmark datasets indicate that the proposed SiamSTM achieves state-of-the-art performance. In future work, we plan to embed the mask prediction end-to-end into this framework to accommodate VOT2020 and VOS tasks.

{\small
\bibliographystyle{ieee_fullname}
\bibliography{egpaper_for_review}
}

\end{document}